\documentclass[10pt,twocolumn,letterpaper]{article}

\usepackage{cvpr}              

\usepackage{graphicx}
\usepackage{amsmath}
\usepackage{amssymb}
\usepackage{booktabs}
\usepackage[export]{adjustbox}
\usepackage{diagbox}

\usepackage[pagebackref,breaklinks,colorlinks]{hyperref}

\usepackage[accsupp]{axessibility} 

\usepackage[capitalize]{cleveref}
\crefname{section}{Sec.}{Secs.}
\Crefname{section}{Section}{Sections}
\Crefname{table}{Table}{Tables}
\crefname{table}{Tab.}{Tabs.}

\def\cvprPaperID{41} 
\def\confName{CVPR}
\def\confYear{2022}

\begin{document}

\title{Event Transformer. A sparse-aware solution for efficient event data processing}


\author{Alberto Sabater $^{1}$ ~~~~~~ Luis Montesano$^{1,2}$ ~~~~~~ Ana C.~Murillo$^{1}$\\
{\small $^{1}$DIIS-I3A, Universidad de Zaragoza, Spain ~~~~~ \small $^{2}$Bitbrain Technologies, Spain.}
}

\maketitle

\begin{abstract}

Event cameras are sensors of great interest for many applications that run in low-resource and challenging environments. They log sparse illumination changes with high temporal resolution and high dynamic range, while they present minimal power consumption.   
However, top-performing methods often ignore specific event-data properties, leading to the development of generic but computationally expensive algorithms.
Efforts toward efficient solutions usually do not achieve top-accuracy results for complex tasks.
This work proposes a novel framework, \textit{Event Transformer} (\textit{EvT})\footnote{Code, trained models and supplementary video can be found in: https://github.com/AlbertoSabater/EventTransformer}, that effectively takes advantage of event-data properties to 
be highly efficient and accurate.
We introduce a new patch-based event representation and a compact transformer-like architecture to process it.
EvT is evaluated on different event-based benchmarks for action and gesture recognition.
Evaluation results show better or comparable accuracy to the state-of-the-art while requiring significantly less computation resources, which makes EvT able to work with minimal latency both on GPU and CPU. 

\end{abstract}


\section{Introduction}

Event cameras are bio-inspired sensors that register changes in intensity at each pixel of the sensor array.
Contrary to traditional cameras, they work in a sparse and asynchronous manner, with an increased High Dynamic Range and high temporal resolution (in the order of microseconds) with minimal power consumption. 
These characteristics have 
pushed the research of many event-based perception tasks such as 
action recognition \cite{bi2020graph, innocenti2021temporal},
body \cite{calabrese2019dhp19, rudnev2021eventhands}
and gaze tracking \cite{angelopoulos2021event},
depth estimation, \cite{gehrig2021combining, wang2021stereo}
or odometry \cite{klenk2021tum, rodriguez2021griffin},
interesting for many applications that involve low-resource environments and challenging motion and lightning conditions, such as AR/VR or autonomous driving.

Processing information from event-based cameras is still an open research problem. 
Top-performing approaches transform event-streams into frame-like representations, throwing away their inherent sparsity, and using heavy processing algorithms such as Convolutional Neural Networks \cite{amir2017low, innocenti2021temporal, baldwin2021time, cannici2020differentiable} or Recurrent Layers \cite{innocenti2021temporal, cannici2020differentiable}.
Other methods that better exploit this sparsity, such as 
PointNet-like Neural Networks \cite{wang2019space}, Graph Neural Networks \cite{bi2020graph, deng2021ev} or Spike Neural Networks \cite{kaiser2020synaptic, shrestha2018slayer},
are more efficient but do not reach the same accuracy. 
As a result, there is a need for methods that put to good use all the potential of event-based cameras for efficiency and low energy consumption while maintaining high performance.

\begin{figure}
\centering
\includegraphics[width=0.95\linewidth]{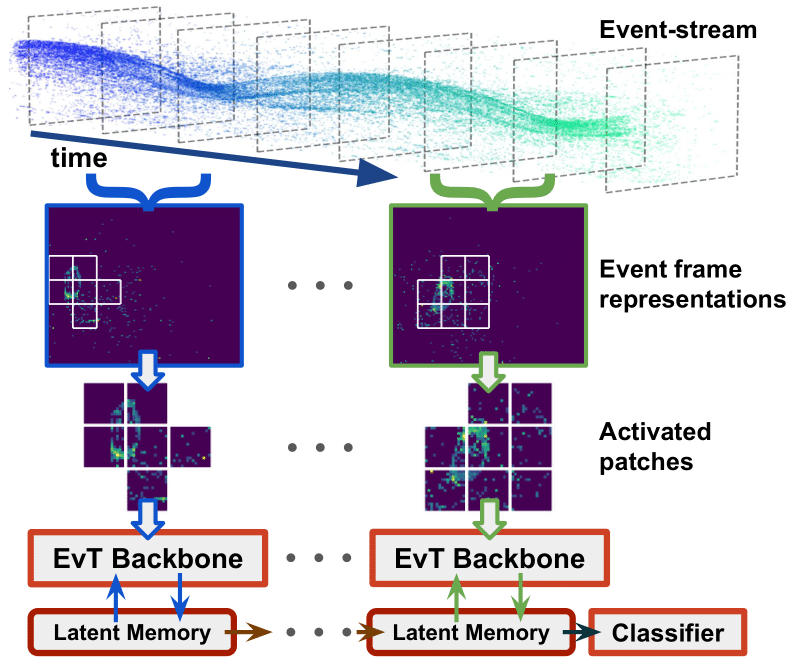}
\caption{\textbf{Framework overview}. 
Areas (\textit{activated patches}) from event-streams with sufficient event information are extracted from event frame representations and processed by the EvT backbone to update a set of \textit{latent memory vectors}.
The latest version of this memory is used for a final event-stream classification.
}
\label{fig:intro}
\end{figure}

This work introduces \textit{Event Transformer} (\textit{EvT}), a novel framework (summarized in Fig. \ref{fig:intro}) 
designed to tackle the event data sparsity to be highly efficient while obtaining top-accuracy results.
We propose:
1) the use of a sparse \textit{patch-based event data representation}, that only accounts for the areas of event-streams with registered information, 
and 
2) a compact transformer-like backbone based on attention mechanisms \cite{vaswani2017attention} that naturally work with this patched information. The later, in contrast to previous frame-based methods, requires minimal computational resources by using a set of \textit{latent memory vectors} to bound the quadratic computational complexity associated to transformer architectures \cite{jaegle2021perceiver, jaegle2021perceiverio}, 
but also, to encode the seen information.

EvT evaluation is run on three public real event data benchmarks of different complexity for long and short event-stream classification (i.e., action and gesture recognition).
The results show that EvT achieves better or comparable results
to the state-of-the-art. More importantly, our approach presents a significant decrease in computational resource requirements with respect to previous works, with the corresponding power consumption savings, making EvT able to work with minimal latency both in GPU and CPU.

\section{Related work}

This section summarizes the most common approaches for event data representation as well as event-based Neural Network architectures that process them. 
It also includes a brief description of different kind of event datasets.

\subsection{Event data representation} 
Event data representations encode the event information related to a time-interval or temporal-window extracted from an event-stream. These representations can be divided in two categories:
\textbf{event-level representations} usually treat the event data as graphs \cite{wang2019space, bi2019graph, bi2020graph, deng2021ev} or point-clouds \cite{sekikawa2019eventnet, Vemprala2021RepresentationLF} with minimal pre-processing and keeping the event data sparsity; 
differently, \textbf{frame-based representations} group incoming events into dense frame-like arrays, ignoring the event data sparsity but easing a later learning process.
Our work is built on the top of frame-based representations, where we find plenty of variations in the literature. 
The \textit{time-surfaces}\cite{lagorce2016hots} build frames encoding the last generated event for each pixel. 
SP-LSTM \cite{nguyen2019real} builds frames where each pixel contains a value related to the existence of an event in a time-window and its polarity.
The \textit{Surfaces of Active Events} \cite{mueggler2015lifetime} builds frames where each pixel contains a measurement of the time between the last observed event and the beginning of the accumulation time.
\textit{Motion-compensated}\cite{rebecq2017real, vidal2018ultimate} generate frames by aligning events according to the camera ego-motion.
\cite{ghosh2019spatiotemporal} binarizes frame representations in the temporal dimension, achieving a better time-resolution.
TBR \cite{innocenti2021temporal} aggregates binarized frame representations into single-bins frames. 
M-LSTM \cite{cannici2020differentiable} uses a grid of LSTMs that processes incoming events at each pixel to create a final 2D representation.


Our work introduces a \textit{patch-based event representation}: we first build a simple frame representation (similar to \cite{ghosh2019spatiotemporal}), and later we divide the resulting frames into a grid of non-overlapping patches, inspired by ideas from Visual Transformers \cite{dosovitskiy2020image}.
Generated patches with not sufficient event information are discarded, while the rest are keep as the final event data representation.
The proposed hybrid solution presents benefits from both the event-level representations, since we can to a certain extent tackle the sparsity of the event data, but also from the robustness of the frame-based representations. 

\subsection{Neural Network architectures for event data}
Deep learning based techniques have shown promising results working with event-camera data. This section discusses the main existing architectures to process different types of event representations, as well as to aggregate the processed information from several time-windows. 
Besides, 
we provide a short overview of Visual Transformers, since they are actually one of the pillars of the architecture proposed in this work to process events.

\paragraph{Architectures for event-representations processing.}
Methods designed to process event-level representations 
process the information within each time-window 
with architectures that take advantage of the event sparsity such as Spike Neural Networks \cite{kaiser2020synaptic, shrestha2018slayer, wu2018spatio}, PointNet-style Networks \cite{wang2019space} or Graph neural Networks \cite{bi2019graph, bi2020graph, deng2021ev}.
Differently, frame-based methods often rely on the use of Convolutional Neural Networks to process the frames built for each time-window \cite{amir2017low, innocenti2021temporal, baldwin2021time, cannici2020differentiable}. 
Sometimes using also Transformers to process long-range spatial dependencies among CNN-generated features \cite{weng2021event}.
These methods process whole frames, even if no events are triggered in large parts of this frame. Despite usually achieving higher accuracy than event-level representations, 
this unnecessary processing makes frame-based approaches 
consume more computational resources than needed.

\paragraph{Architectures to aggregate several time-windows data.} 
Solutions aimed to analyze long event-streams divide them into different time-windows that are represented and processed with the methods previously described, and then aggregated to perform the final visual recognition task.
This aggregation is performed differently in the related work, including the use of Recurrent Networks \cite{innocenti2021temporal, weng2021event},
CNNs \cite{amir2017low, innocenti2021temporal},
temporal buffers \cite{baldwin2021time, deng2021ev},
or voting between the intermediate results of each time-window \cite{innocenti2021temporal}.

Depending on the aggregation strategy, we consider that an event-processing algorithm is able to perform \textbf{online inference} if it can evaluate the information within each time-window incrementally, as it is generated, and then perform the final visual recognition with minimal latency, as opposed to the processing of all the captured information in a large batch. 
Our approach performs online inference by updating incrementally a set of latent memory vectors with simple addition operations, and processing the resulting vectors with a simple classifier.

\paragraph{Visual Transformers.}

Transformer architectures, initially introduced for Natural Language Processing \cite{vaswani2017attention}, have recently gained popularity for Visual Recognition tasks.
Vision transformers, work on the features generated by CNNs \cite{carion2020end} from RGB images or, more commonly, are directly applied to patches extracted from the input images \cite{dosovitskiy2020image, liu2021swin}.
When working on video data, 
some methods \cite{arnab2021vivit, liu2021video} process patches from different video frames together
or aggregate the temporal information with LSTM layers \cite{huang2020multimodal}.
Of special relevance for this work is Perceiver \cite{jaegle2021perceiver, jaegle2021perceiverio}, a transformer that uses latent vectors to process the input data and  bound the quadratic complexity of transformers.

In this work, we use a Transformer-like architecture to analyze sets of activated patches of variable length, extracted from time-windows within an event-stream.
Inspired by \cite{jaegle2021perceiver, jaegle2021perceiverio}, we use
latent vectors for this processing, but differently, we also refine them incrementally with the information extracted from different time-windows.

\subsection{Event dataset recordings}\label{sec:datasets}
Despite their promising applicability, there are still not many large-scale public datasets recorded with these cameras in real scenarios.
Therefore, some methods seek for the translation of RGB datasets to their event-based counterpart. 
Earlier event-based solutions \cite{orchard2015converting, serrano2013128, li2017cifar10, bi2020graph, hu2016dvs}
display RGB data in a LCD monitor that they record with an event-camera. 
More recent works introduce the use of learning-based emulators \cite{nehvi2021differentiable, hu2021v2e, gehrig2020video} to generate event data. 
Still, these translated datasets cannot fully mimic the event-data nature and introduce certain artifacts, specially on their sparsity and latency. This happens because events are triggered by unrealistic lightning conditions and are frequently dependent on the fixed low frame-rate of a monitor or the movement of a camera over static images.
In order to have a more reliable evaluation setup, we focus our experimentation on datasets recorded with event-cameras on real scenarios.
More specifically, we train and evaluate EvT in the classification of both short \cite{bi2019graph} and long \cite{amir2017low, vasudevan2021sl} event-streams related to action and gesture recognition. 

\section{Event Transformer framework}

\begin{figure*}[t]
    \centering
    
    \subfloat[Backbone overview]
    {\label{fig:model_details}\includegraphics[width=0.72\linewidth]{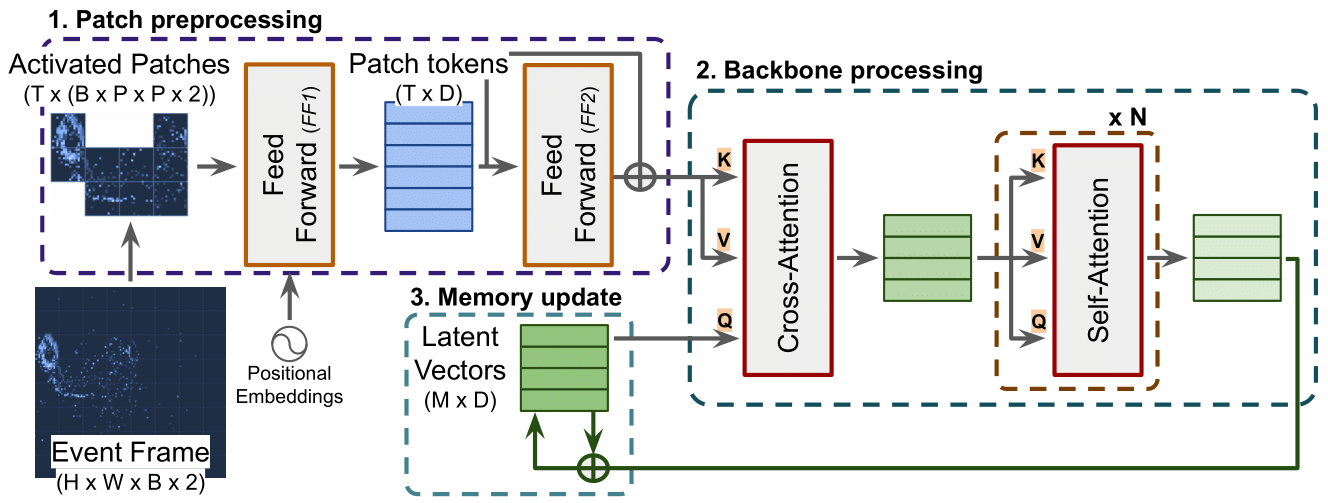}}
    \hfill\vline\hfill
    \subfloat[Classification {architecture}]
    {\label{fig:classifier_details}
    \includegraphics[width=0.12\linewidth]{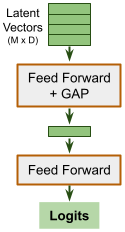}
    }
    \hfill\vline\hfill
    \subfloat[Cross and Self-Attention architecture]
    {\label{fig:block_details}
    \includegraphics[width=0.115\linewidth]{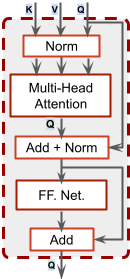}
    }
    \caption{Event Transformer (EvT) overview. 
    (a) For each new event frame built from an input event-stream, \textit{Activated Patches} are detected and pre-processed to build patch tokens. These tokens are processed by our backbone, and the resulting output is used to update a set of latent vectors. These latent vectors encode all the information received so far.
    (b) The final version of latent vectors is used to perform the final event-stream classification.
    (c) Architecture shared by the Cross and Self-Attention modules.
    }
    \label{fig:model_overview}
\end{figure*}

Different to traditional RGB cameras, event-cameras log the captured visual information in a sparse and asynchronous manner.
Each time an intensity change is detected, the camera triggers an event $e=\{x,y,t,p\}$ defined by its location ($x,y$) within the space of the sensor grid ($H \times W$), the timestamp $t$ of the event (in the order of $\mu s$) and its polarity $p$ (either positive or negative change).

This work proposes a novel solution (overview in Fig.~\ref{fig:intro}) for efficient event data processing. 
Our framework introduces a \textbf{patch-based event data representation} that extracts the areas of intermediate frame representations with sufficient logged event information.
The resulting information is then processed by our proposed \textbf{Event Transformer (EvT)}, a compact Neural Network that
uses a set of latent memory vectors to process the incoming event data as well as to encode the information seen so far.
The final version of these latent vectors is processed to perform the final visual recognition task, event-stream classification in our case.
Next subsections detail the proposed patch-based event data representation and Event Transformer processing, and how the memory vectors are used for the classification of a stream of events.

\subsection{Patch-based event data representation}\label{sec:event_representation}

Similar to previous frame-based methods \cite{innocenti2021temporal, amir2017low, baldwin2021time}, we aggregate the event data $\varepsilon = \{e^1, e^2, e^3, ...\} \mid e_t^i \epsilon \Delta t$ generated during a time-window $\Delta t$ into a frame-like representation $F^{H \times W \times B \times 2}$.
Each location $(x,y) \mid y \epsilon H, x \epsilon W$ in $F$ is represented with two histogram-like vectors of $B$ bins, one for each polarity $p \epsilon \{0,1\}$. Each histogram discretizes $\Delta t$ in each bin and counts the number of positive or negative events occurring in the corresponding period $\Delta t / B$.
Final representations are transformed as $F' = log(F + 1)$ to smooth extreme high values in highly activated areas.

Frame representations, are then split into non-overlapping patches of size $P\times P$.  
Then we set each patch as activated if it contains at least $m$ percent of non-zero elements, i.e., if at least a $m$ percent of the pixels $(x,y)$ within the patch have registered events.
Activated patches are kept for further processing by EvT, while the non-activated patches are discarded, reducing significantly the following computation cost and implicitly the ambient noise. 
If the amount of activated patches is below a threshold $n$, i.e., there is not enough visual information to be processed, we expand the time-window $\Delta t$  and increase the event set $\varepsilon$ with the new incoming events, recompute the frame representation and extract the activated patches. This last step is repeated until we get at least $n$ activated patches. As a final step, we flatten the $T$ activated patches to create tokens of size ${(P^2 \times B \times 2)}$, input of the transformer backbone detailed in the next section. 

This patch-based event representation is therefore designed to take advantage from different event-data properties for a later more efficient event-data processing.
Of special relevance, the \textbf{spatio-temporal event sparsity} is addressed by dropping non-informative patches, and the \textbf{low latency} of this data is addressed by using short time-windows (whose length is adapted to the events density), which are also binarized for a finer representation.

\subsection{Event Transformer}

Transformers are a natural way to process the patch-based representation we propose. 
Different to other architectures, they are able to ingest lists of tokens of variable length 
that are processed with attention mechanisms.
The later, different to convolutions, focus on the whole input data (structuring it as a Query ($Q$), Key ($K$) and Value ($V$)) to capture both local and long-rage token dependencies.

The processing core of our work, Event Transformer (EvT), is motivated by these ideas.
This backbone processes, with attention mechanisms and a set of $M$ learned latent vectors, lists of patch tokens (whose length varies depending on the event-data sparsity). These latent vectors serve also as a memory that is incrementally refined with the processing of activated patches calculated, as described in Section \ref{sec:event_representation}, from consecutive time-windows within the event-stream.
The final version of the refined latent vectors is processed with a simple classifier to perform the final event-stream classification.
The whole process (detailed in Figure \ref{fig:model_overview}) is divided in the following steps:


\paragraph{Patch pre-processing.}
Each one of the $T$ input activated patches is mapped to a vector of dimensionality $D$. This vector length $D$ is constant along all the network.
This transformation (\textit{FF1}) consists of an initial single-layer Feed Forward Network (FF), the concatenation of 2D-aware positional embeddings, and a last single-layer FF Network.
The use of positional embeddings to augment the patch information is required since Transformers, unlike CNNs, cannot implicitly know the locality of the input data.  
Before being processed by the core of our backbone, transformed patches, i.e., patch tokens, undergo another transformation (\textit{FF2}) composed of a double-layer Feed-Forward Network and a skip-connection to achieve a finer representation while preserving the information about its locality.

\paragraph{Backbone processing.}
The core of our backbone is composed by a single Cross-Attention and $N$ Self-Attention modules that share the same architecture (detailed in Fig. \ref{fig:block_details}). 
Similar to previous transformer related works \cite{vaswani2017attention, jaegle2021perceiver, arnab2021vivit}, it is composed of a Multi-Head Attention layer \cite{vaswani2017attention}, normalization layers, skip connections and Feed Forward layers.
Our backbone first processes the latent memory vectors $Q$ on the basis of the patch token information $K$-$V$ (Cross-Attention)
and resulting vectors are then refined $Q$-$K$-$V$ with no external information (Self-Attention). 

\paragraph{Memory update.}
Once the patch tokens $T$ and latent vectors have been processed by the backbone, the resulting latent vectors in this iteration are combined with the existing memory latent vectors with a simple sum operation. This augmented version of latent vectors encodes a broader spatio-temporal information and will be used to process the activated patches from the following time-window.

\paragraph{Classification output.}
The final output of our proposed framework is obtained by processing the latest version of the latent memory vectors, that contain the key spatio-temporal information of the event-stream seen so far. 
In our case, we perform a multi-class classification by simply processing the latent vectors with two Feed Forward Layers and Global Average Pooling (GAP), as detailed in Fig. \ref{fig:classifier_details}.

\section{Experiments}

This section includes the implementation and training details of the proposed Event Transformer (EvT), and the experimental validation. We evaluate EvT in two tasks (long and short event-stream classification for action and gesture recognition), analyze its efficiency, and justify the EvT design choices with a thorough ablation study.

\subsection{Implementation and training details}\label{sec:implementation}

\paragraph{Patch-based event representation.}
We set a common patch size of $6 \times 6$ pixels, but specific values for time-window $\Delta t$ and number of bins $B$ are specific for each dataset and defined in the following subsections.
For all cases, to consider a patch as \textit{activated} we set $m=7.5$, i.e., we require a $7.5\%$ of the pixels  within the patch to log events. 
Besides, we set $n=16$, i.e., 
an event frame must have at least 16 active patches to continue the processing. As previously detailed, if the frame does not have enough ($\geq 
n$) active patches we increase the time-window covered by the frame and repeat the search of active patches.

\paragraph{Event Transformer.}
The dimensionality $D$ of the latent vectors and the pre-processed patch tokens is set to $128$.
The latent memory is composed of $96$ latent vectors randomly initialized at the beginning of the training with a Normal distribution of mean $0.0$ and deviation $0.2$.
Similarly, the positional encodings are initialized with 16 bands of 2D Fourier Features \cite{tancik2020fourier} (dimensionality of $\frac{H}{P} \times \frac{W}{P} \times 64$, being $H$ and $W$ the specific sensor height and width from each dataset). 
Both the latent vectors and positional encodings are learned as the rest of the parameters of the network during training.
Patch tokens are processed by a single Multi-Head Cross-Attention layer and $2$ Multi-Head Self-Attention layers, all of them using 4 attention heads.

\paragraph{Training details.}
The whole framework is optimized with a Negative Log Likelihood and AdamW \cite{loshchilov2017decoupled} optimizer, in a single NVIDIA Tesla V100.
The initial learning rate is set as $1e-3$, and we use Stochastic Weight Averaging \cite{izmailov2018averaging} and gradient clipping.
Due to computation resource constraints, we perform the ablation experiments with a batch size of $64$ and we reduce the learning rate by a factor of 0.5 after 10 epochs with no loss reduction.
Differently, the benchmark results are obtained using a batch size of $128$ and a \textit{1cycle learning rate} policy \cite{smith2019super} for 240 epochs.
As for the data augmentation, we use spatial and temporal random cropping, dropout, drop token, and we repeat each sample within the training batch twice with different augmentations.

\subsection{Evaluation}

The performance of the proposed Event Transformer (EvT) is evaluated in two scenarios (visualized in the supplementary video) of real event-camera recordings that represent different use cases.
First, we evaluate EvT to classify long event-streams, where the analysis and aggregation of the information from different sections (temporal-windows) of the stream is key. 
Second, we demonstrate that our solution is also suitable to classify short event-streams, where the recorded target has little motion and less video information is available.
Note that, since the methods we compare EvT with are not always designed for both long and short event-stream classification, we cannot show their accuracy in all the evaluated datasets.
We discuss the EvT efficiency on each classification task with respect to other solutions 
and we show the ability of EvT to take advantage of the event data sparsity with an efficiency analysis on different kinds of datasets.

\subsubsection{Long event-stream classification}\label{sec:long_clf}

EvT is evaluated in two benchmarks for long event-stream classification.
The \textbf{DVS128 Gesture Dataset} \cite{amir2017low} is composed of 1342 event-streams capturing 10 different human gestures (plus an extra category for random movements) and recorded with 29 different subjects under three different illumination conditions.
The \textbf{SL-Animals-DVS Dataset} \cite{vasudevan2021sl} is composed of 1121 event-streams capturing 19 different sign language gestures, executed by 58 different subjects, under different illumination conditions.
Recordings from these two datasets last $1$-$6$ s.
Recordings are split, as detailed in Section \ref{sec:ablation}, into time-windows $\Delta t$ of 24ms for the DVS128 dataset and 48ms for the SL-Animals-DVS.
Similar to previous works \cite{bi2020graph, innocenti2021temporal, wang2019space, baldwin2021time}, we build intermediate event representations for each time-window extracted from the event-stream. Then, we process and aggregate them iteratively for the final event-stream classification.

\begin{table}[h]
\footnotesize
    \centering
    \begin{tabular}{|l|c|c|c|} \hline 
        \textbf{Model} & \textbf{10 Classes} & \textbf{11 Classes} & \textbf{Online}\\ \hline \hline
        RG-CNN \cite{bi2020graph} & N/A & 97.2 & x \\ \hline
        3D-CNN + Voting \cite{innocenti2021temporal} & 99.58 & 99.62 & x \\ \hline \hline
        CNN \cite{amir2017low} & 96.49 & 94.59 & \checkmark \\ \hline 
        Space-time clouds \cite{wang2019space} & 97.08 & 95.32 & \checkmark \\ \hline
        CNN + LSTM \cite{innocenti2021temporal} & 97.5 & \textbf{97.53} & \checkmark \\ \hline
        TORE 
        \cite{baldwin2021time} & N/A & 96.2 & \checkmark \\ \hline
        \textbf{EvT (Ours)} & \textbf{98.46} & 96.20 & \checkmark \\ \hline
    \end{tabular}
    \caption{Classification Accuracy in DVS128 Gesture Dataset (long event-streams). N/A = Not Available at the source reference}
    \label{tab:DVS128}
\end{table}

\begin{table}[!h]
    \centering
    \footnotesize
    \begin{tabular}{|l|c|c|} \hline 
        \textbf{Model} & \textbf{3 Sets} & \textbf{4 Sets} \\ \hline \hline
        SLAYER \cite{vasudevan2020introduction} & 78.03 & 60.09 \\ \hline
        STBP \cite{vasudevan2020introduction} & 71.45 & 56.20 \\ \hline
        DECOLLE \cite{kaiser2020synaptic} & 77.6 & 70.6 \\ \hline
        TORE 
        \cite{baldwin2021time} & N/A & 85.1 \\ \hline
        \textbf{EvT (Ours)} & \textbf{87.45} & \textbf{88.12} \\ \hline
    \end{tabular}
    \caption{Classification Accuracy in SL-Animals-DVS (long event-streams). N/A = Not Available at the source reference}
    \label{tab:SL-Anim}
\end{table}

Table \ref{tab:DVS128} shows the accuracy of top-performing models in the DVS128 Dataset, with and without including the extra additional distractor of random movements (11 and 10 classes classification respectively). The column \textit{Online} highlights the ability of each model to perform online inference, 
i.e., incremental processing of the event data and classification with low latency.
Similarly, Table \ref{tab:SL-Anim} shows the accuracy of top-performing methods evaluated on the SL-Animals-DVS Dataset, a more demanding benchmark with lower state-of-the-art accuracy. \textit{3 Sets} results exclude the samples recorded indoor with artificial lighting from a neon light source,
since they include noise related to the reflection of clothing and the flickering of the fluorescent lamps. \textit{4 Sets} evaluates all the samples within the dataset. 

Results from Table \ref{tab:DVS128} show how our approach obtains comparable results to state-of-the-art. 
Only \cite{innocenti2021temporal} is more accurate than EvT but it uses offline inference and CNNs, which are computationally more expensive but have a good inductive bias, useful when training with small datasets like DVS128 and with random movements (as is the case of 11 Classes).
As for the most challenging SL-Animals Dataset, EvT achieves a new state-of-the-art. Interestingly, our solution presents higher robustness to different lightning conditions. Note how the performance does not decrease when using the \textit{4 Sets} instead of 3. Instead, it is able to take advantage of larger training set and achieves better accuracy.

Regarding the \textbf{model efficiency}, there is not available information to directly compare our approach to the other methods reported in these two datasets.  
However, the best methods in Tables \ref{tab:DVS128} and \ref{tab:SL-Anim}
use CNN-based models \cite{amir2017low, innocenti2021temporal,  baldwin2021time} to process frame-event representations and/or complex aggregation architectures such as Recurrent Layers \cite{innocenti2021temporal} or CNNs \cite{bi2020graph} to process intermediate time-window results. 
Our approach processes event representations with minimal cost compared to other approaches (see Table \ref{tab:flops} for a similar comparison in the next dataset), and incrementally aggregates intermediate results on the latent vectors used for the final classification with negligible cost.
Given this, EvT is able to perform online inference, while being significantly more resource-efficient than the related methods for long event-stream classification.

\subsubsection{Short event-stream classification} \label{sec:short_clf}

This experiment tackles the problem of classifying event-streams of scenes with low motion, typically shorter than those of the previous subsection. 
This experiment is ran on the \textbf{ASL-DVS Dataset} \cite{bi2019graph}. It contains 100,800 event-streams capturing 24 letter signs from the American Sign Language, performed by 5 different subjects. Each sample lasts about 100 ms. The dataset is randomly split with 80\% of the data for training and the remaining 20\% for testing, as proposed by the authors.
As many other methods that tackle this problem \cite{bi2019graph, cannici2020differentiable, deng2021ev, baldwin2021time}, we treat it as a simple instance classification problem where a single event representation (from a single time-window $\Delta t$) is built from the whole event-stream.

To perform this short event-stream classification, we represent each event-stream with a single frame representation that encodes 100 ms of the sample ($\Delta t = 100 ms, B=2$). 
Then, the final classification is performed by processing the generated activated patches with the EvT backbone and computing the logits from the output latent vectors, with no need to perform any memory update. 
Table \ref{tab:asl_bench} shows that EvT is able to get excellent performance.

\begin{table}[h]
    \centering
    \footnotesize
    \begin{tabular}{|l|c|c|} \hline 
        \textbf{Model} & \textbf{Accuracy} \\ \hline \hline
        RG-CNN \cite{bi2019graph} & 90.1 \\ \hline
        EV-VGCNN \cite{deng2021ev} & 98.3 \\ \hline
        M-LSTM 
        \cite{cannici2020differentiable} & 99.73 \\ \hline
        TORE 
        \cite{baldwin2021time} & 99.6 \\ \hline
        \textbf{EvT (Ours)} & \textbf{99.93} \\ \hline
    \end{tabular}
    \caption{Classification Accuracy in ASL-DVS (short event-streams)}
    \label{tab:asl_bench}
\end{table}

Although this particular dataset is already solved satisfactorily by prior work, we find that EvT perform with lower computational resources than them.
Unfortunately, there are no \textbf{model efficiency} statistics published from prior work on real short event-stream benchmarks. However, several works provide them on the simulated dataset \textbf{N-Caltech-101} \cite{orchard2015converting}, so we use this dataset for the following analysis. This dataset is the event counterpart of the RGB images from Caltech-101 \cite{fei2004learning}. 
It contains short event recordings of RGB images displayed on a LCD monitor.
Table \ref{tab:flops} shows the model complexity (measured in FLOPs and number of model parameters, which is directly related to energy consumption) required by different methods evaluated on the N-Caltech-101 dataset. 
EvT presents much lower requirements than frame-based methods, that run heavy computations. 
More importantly, EvT also presents lower computational requirements than point-based methods, that are also designed to tackle the sparsity of the event data, but achieve less computational savings than our approach. 
%
%
%
Note that since EvT is designed with real-event processing in mind, it does not improve the state-of-the-art accuracy on the N-Caltech-101 dataset. 
Therefore, this experiment only intends to present a common setup to measure efficiency (not accuracy).
Previous Table \ref{tab:asl_bench} already showed an accuracy comparison with these short-stream classification approaches when using real event camera data.

\begin{table}[h]
\small
    \centering
    \begin{tabular}{|l|l|c|c|c|} \hline 
        \textbf{Model} & \textbf{Type} & \textbf{(G)FLOPs} & \textbf{\#Params.} \\ \hline 
        RG-CNN \cite{bi2019graph} & Point-based & 0.79  & N/A \\ \hline 
        EV-VGCNN \cite{deng2021ev} & Point-based & 0.70 & 0.84 M \\ \hline  
        M-LSTM \cite{cannici2020differentiable} & Frame-based & 4.82  & 21.43 M \\ \hline 
        \textbf{EvT (Ours)} & \textbf{Patch-based} & \textbf{0.20} & \textbf{0.48 M} \\ \hline    
    \end{tabular}
    \caption{Average FLOPs for all validation samples in N-Caltech-101 and number of parameters per model.   
    N/A = Not Available at the source reference
    }
    \label{tab:flops}
\end{table}

\subsubsection{Event sparsity and model efficiency analysis.}

In the previous Section, we have compared the computational cost and model size of EvT with prior work in the non-real N-Caltech-101 dataset (Table \ref{tab:flops}). 
We now provide a deeper analysis of the computational cost of EvT showing, on four datasets, how it takes advantage of event data sparsity and a compact network design to improve its efficiency. 

The computational cost of EvT is determined by the Cross-Attention Layer, which has $O(T \times M)$, where $T$ stands for the amount of activated patches and $M$ for the amount of latent memory vectors. 
Therefore, the less activated patches there are, the less resources EvT needs. But also, in the worst case, when many activated patches are found ($T \gg M$), 
the latent vectors prevent EvT from incurring a quadratic cost.
Note that the Self-Attention layers have then a reduced cost of $O(M^2)$ instead of $O(T^2)$.

Table \ref{tab:time_perf} shows the average computational cost (time and FLOPs) per time-window $\Delta t$ of EvT, calculated for the different datasets of our previous experimentation. 
As observed, short event-stream datasets (N-Caltech-101 and ASL) generate many patches ($T \gg M$) since they are recorded with a bigger sensor ($240 \times 180$ pixels) and, in the case of N-Caltech-101, because of its synthetic nature.
Otherwise, short event-stream datasets (Sl-Animals and DVS128) generate less patches ($T < M$) since they use a smaller sensor ($128 \times 128$ pixels) and shorter time-windows.
Although the amount of activated patches $T$ of the first group has an order of magnitude more than the second, we do not observe this trend in its computational cost (both in time and FLOPs), that grows smoothly with $T$.

It is important to remark that, in all cases, EvT processes the activated patches in a significantly shorter time-span than the corresponding time-window $\Delta t$, both in GPU and CPU. This highlights the ability of our method to do inference with minimal latency, being capable of processing the event data before the following batch (i.e., time-window data) is generated, therefore facilitating an online inference, even in low resource environments.




\begin{table}[h]
\footnotesize
    \centering
    \begin{tabular}{|l|p{1.2cm}|p{0.4cm}||p{1.5cm}|p{0.9cm}|} \hline 
        \textbf{Dataset} & \textbf{Activated Patches $T$} & \textbf{$\Delta t$ (ms)} & \textbf{Time per $\Delta t$ (GPU/CPU)} & \textbf{(G)FLOPs per $\Delta t$} \\ \hline \hline
        N-Caltech-101   & 532    & 100 & 4 / 16 ms  & 0.20 \\ \hline
        ASL             & 263    & 100 & 4 / 9 ms   & 0.13 \\ \hline \hline
        SL-Animals      & 80     & 48  & 3 / 5 ms   & 0.09 \\ \hline
        DVS128          & 45     & 24  & 2 / 4 ms   & 0.08 \\ \hline
        \multicolumn{5}{@{}l}{\footnotesize{GPU: NVIDIA GeForce RTX 2080 Ti. CPU: Intel Core i7-9700K}}\\
        \multicolumn{5}{@{}l}{\footnotesize{ $\Delta t$: time-window length used to build the initial frame representations}}\\
    \end{tabular}
    \caption{EvT efficiency analysis: execution time and FLOPs per $\Delta t$. Average results for all validation samples in each dataset.}
    \label{tab:time_perf}
\end{table}

\subsection{Ablation study}\label{sec:ablation}

This subsection analyzes the influence of key components in our framework, i.e., event data representation and transformer hyperparameters, to justify our design choices. This evaluation has been performed with both the DVS128 Gesture (10 classes) and the SL-Animals-DVS (4 sets) datasets. 
Note that, as detailed in Section \ref{sec:implementation}, due to computation constraints the following experiments are performed with slightly different training conditions than the ones previously reported.
Therefore, following results properly compare each component of our framework, but the reported accuracy might be inferior to the benchmark ones.

\begin{figure}[!ht]
    \centering
    \subfloat[Time-window $\Delta t$ (ms)]
    {\label{fig:chunk_len}\includegraphics[width=0.48\linewidth]{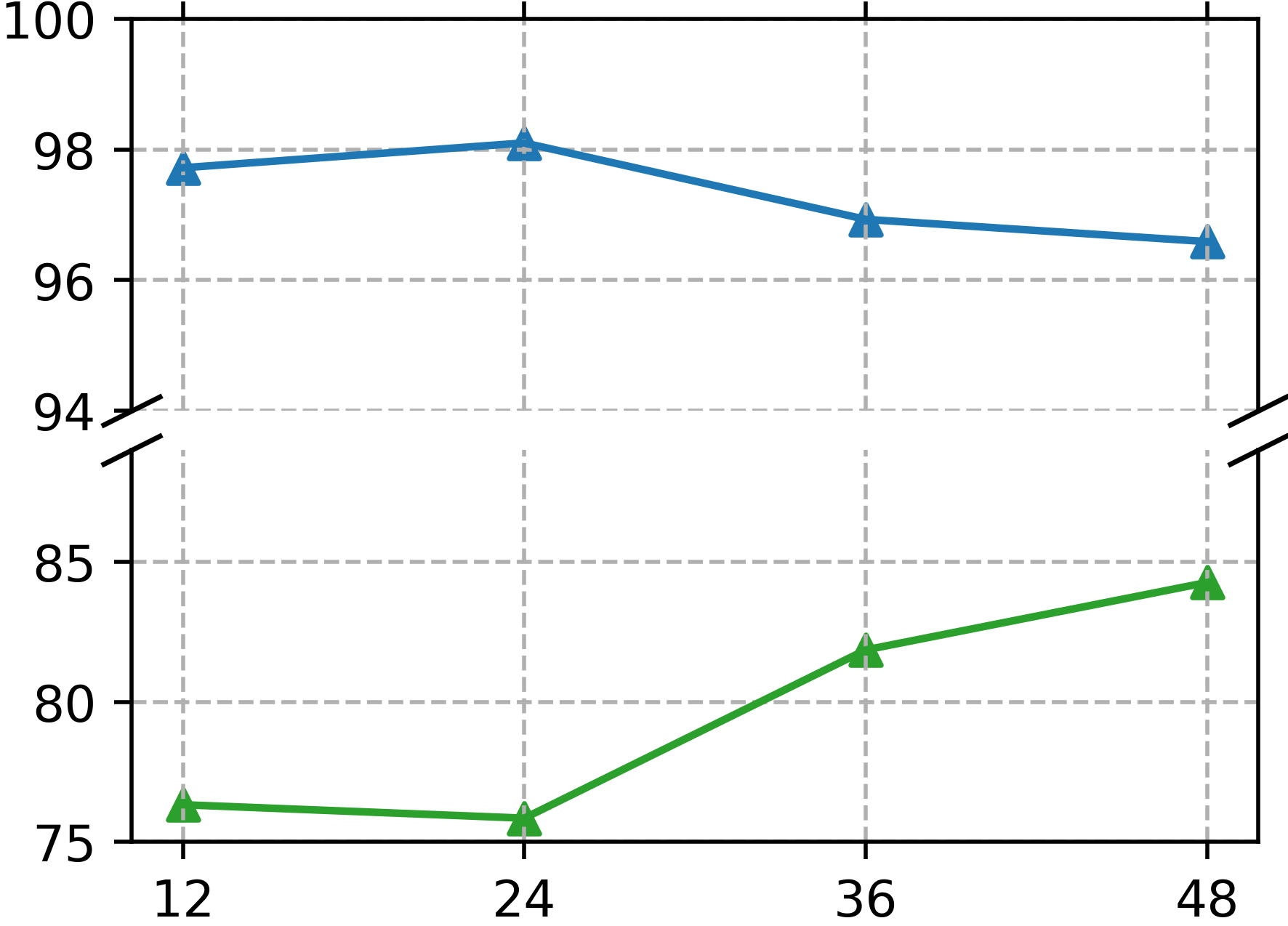}}\hspace{0.15cm}
    \subfloat[Bins $B$]
    {\label{fig:bins}\includegraphics[width=0.48\linewidth]{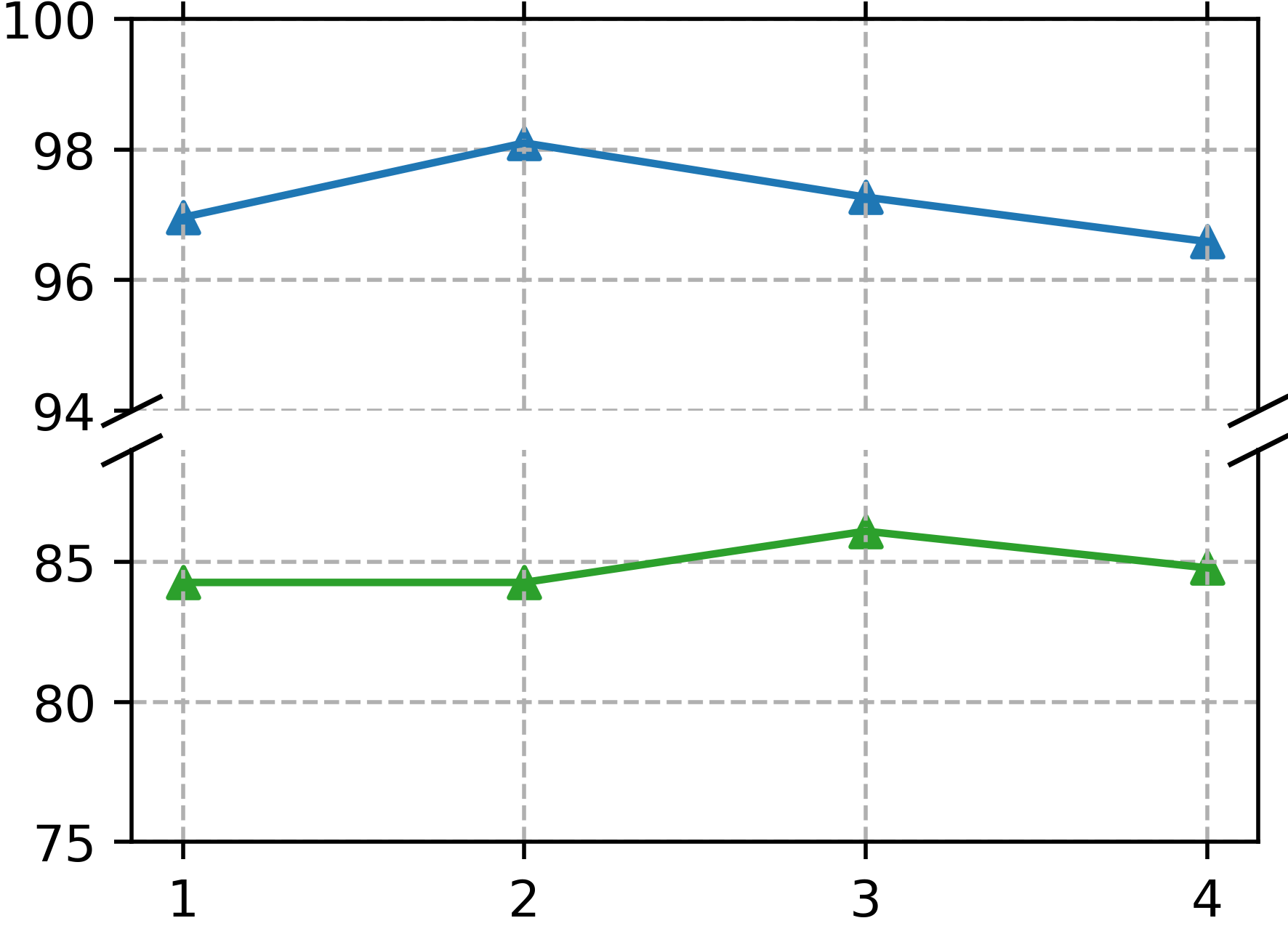}}
    \caption{EvT accuracy with different time-windows $\Delta t$ and bins $B$ for the DVS128 Dataset - 10 classes (top blue line) and for the SL-Animals-Dataset - 4 Sets (bottom green line).}
    
\end{figure}

\paragraph{Event representation hyperparameters.} 
The key components of our event representation are the time-window $\Delta t$ used to aggregate events into a patch representations, and the number of bins $B$ used to improve their time-resolution. 
Figure \ref{fig:chunk_len} shows the EvT accuracy using different time-window lengths $\Delta t$ (and bins $B=2$). As observed, the best $\Delta t$ value is dependent of the evaluated dataset, requiring longer time-windows for SL-Animals-DVS (48 ms) than for DVS128 (24 ms); these time-window values are now set as default for these datasets.
Intuitively, shorter time-windows are required to model motions that are executed at higher speeds, and longer ones fit better slower motions that register less event information.
Figure \ref{fig:bins} shows that using more bins $B$ (3) is beneficial when using longer time-windows (SL-Animals-DVS) and using less bins $B$ (2) helps with shorter time-windows (DVS128); these bins values are set now as default for these datasets.
As observed, the use of more bins, helps us to use finer time-resolutions and therefore improve the final classification accuracy. However, too many bins represent very short time-intervals that do not contain representative event-information.

\begin{figure}[h]
    \centering
    \includegraphics[width=0.55\linewidth]{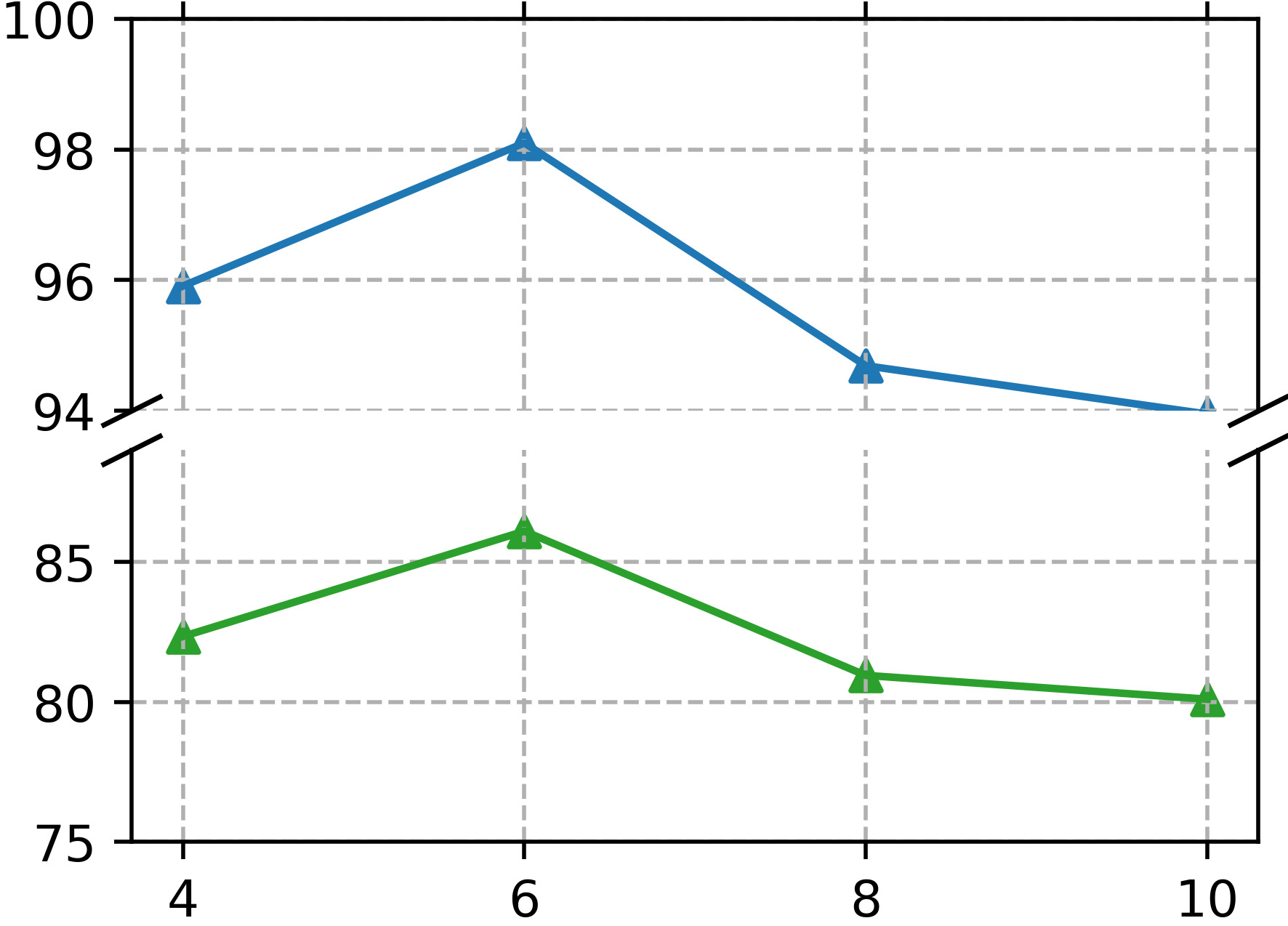}
    \captionof{figure}{EvT Accuracy with different patch size value for the DVS128 Dataset - 10 classes (top blue line) and for the SL-Animals-Dataset - 4 Sets (bottom green line).}
    \label{fig:patch_size}
\end{figure}

Regarding the generation of active patches from event representations, the patch size $P$ is the most relevant hyperparameter, defining the granularity of the information we are working with.
As observed in Figure \ref{fig:patch_size},
smaller patch sizes generate more active patches, making EvT to process more information, while bigger patches speed up the EvT inference, but provide less spatial information.

Besides the patch size, the generation of active patches also depends on the minimum amount of pixels $m$ with logged events needed to activate a patch, and the minimum amount of patches $n$ required to start the EvT backbone processing. 
Table \ref{tab:n_m} shows how filtering
out patches with less event information improves the final accuracy, mainly because it removes the patches that contain noisy event information, not related with the motion in the scene itself.
Additionally, setting a minimum amount of activated patches helps to avoid the processing of time-intervals with no motion and therefore with few activated patches.

\begin{table}[ht]
\small
    \centering
        \begin{tabular}{|p{0.4cm}|p{0.6cm}|p{0.6cm}|p{0.6cm}|}\hline 
            \diagbox[width=0.8cm]{\textbf{$n$}}{\textbf{$m$}} & \textbf{$5\%$} & \textbf{$7.5\%$} & \textbf{$10\%$} \\ \hline 
            \textbf{$8$}  & 97.6 84.6 & 97.9 83.9 & 96.6 83.3 \\ \hline 
            \textbf{$16$} & 96.9 85.9 & 98.1 86.1 & 98.0 84.3 \\ \hline 
            \textbf{$24$} & 97.2 84.3 & 97.2 82.4 & 96.9 82.0 \\ \hline 
        \end{tabular}
      \captionof{table}{EvT accuracy with different combinations of $m$: min. pixels per patch and $n$: min. patches per $\Delta t$. Top cell value: DVS128 (10 classes). Bottom cell value: SL-Animals (4 Sets).}
      \label{tab:n_m}
\end{table}

\begin{figure}[t]
    \centering
    \subfloat[Number of Latent Vectors]
    {\label{fig:num_lat_vec}\includegraphics[width=0.48\linewidth]{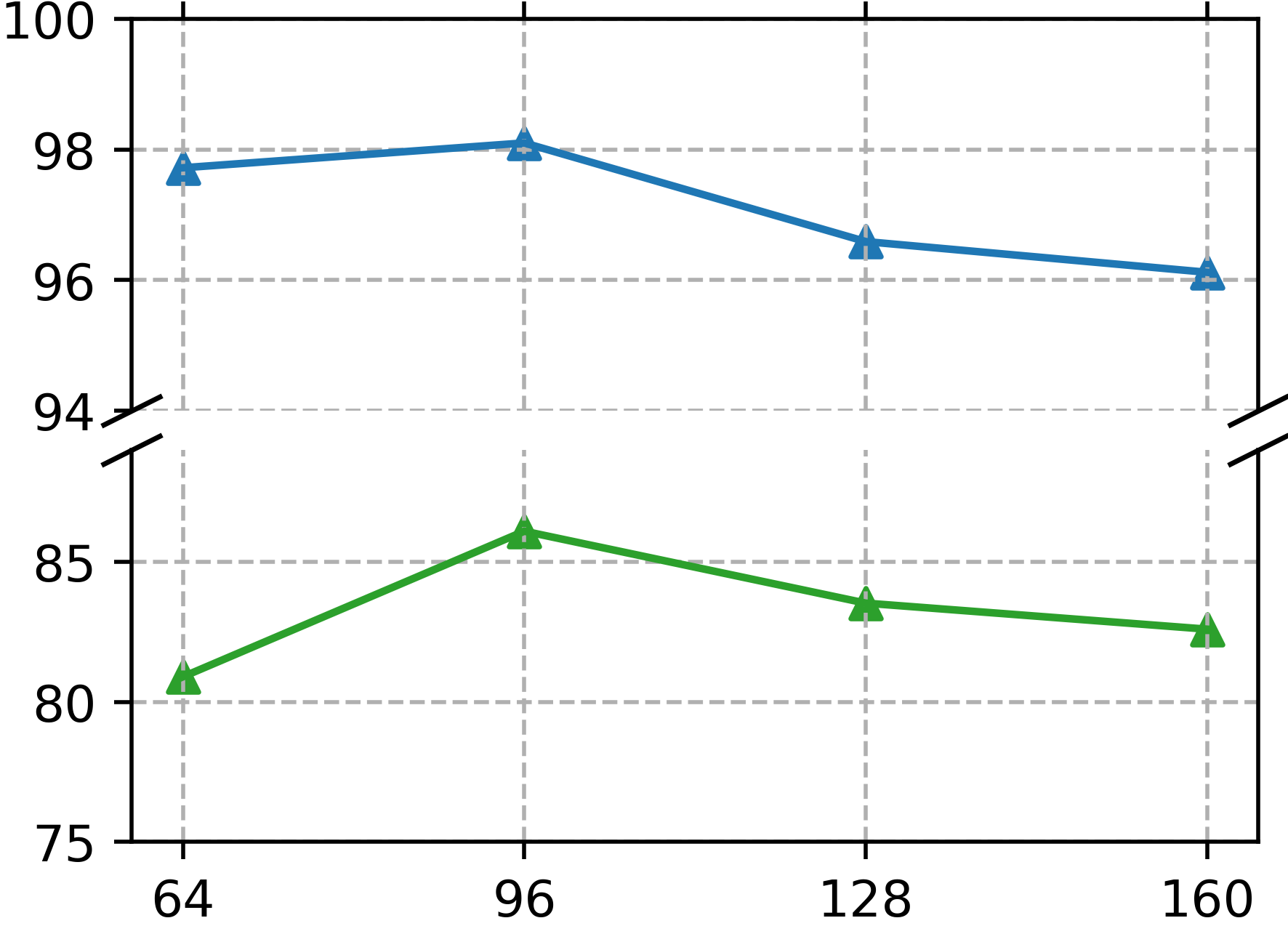}}\hspace{0.15cm}
    \subfloat[Embedding dimension]
    {\label{fig:embed_dim}\includegraphics[width=0.48\linewidth]{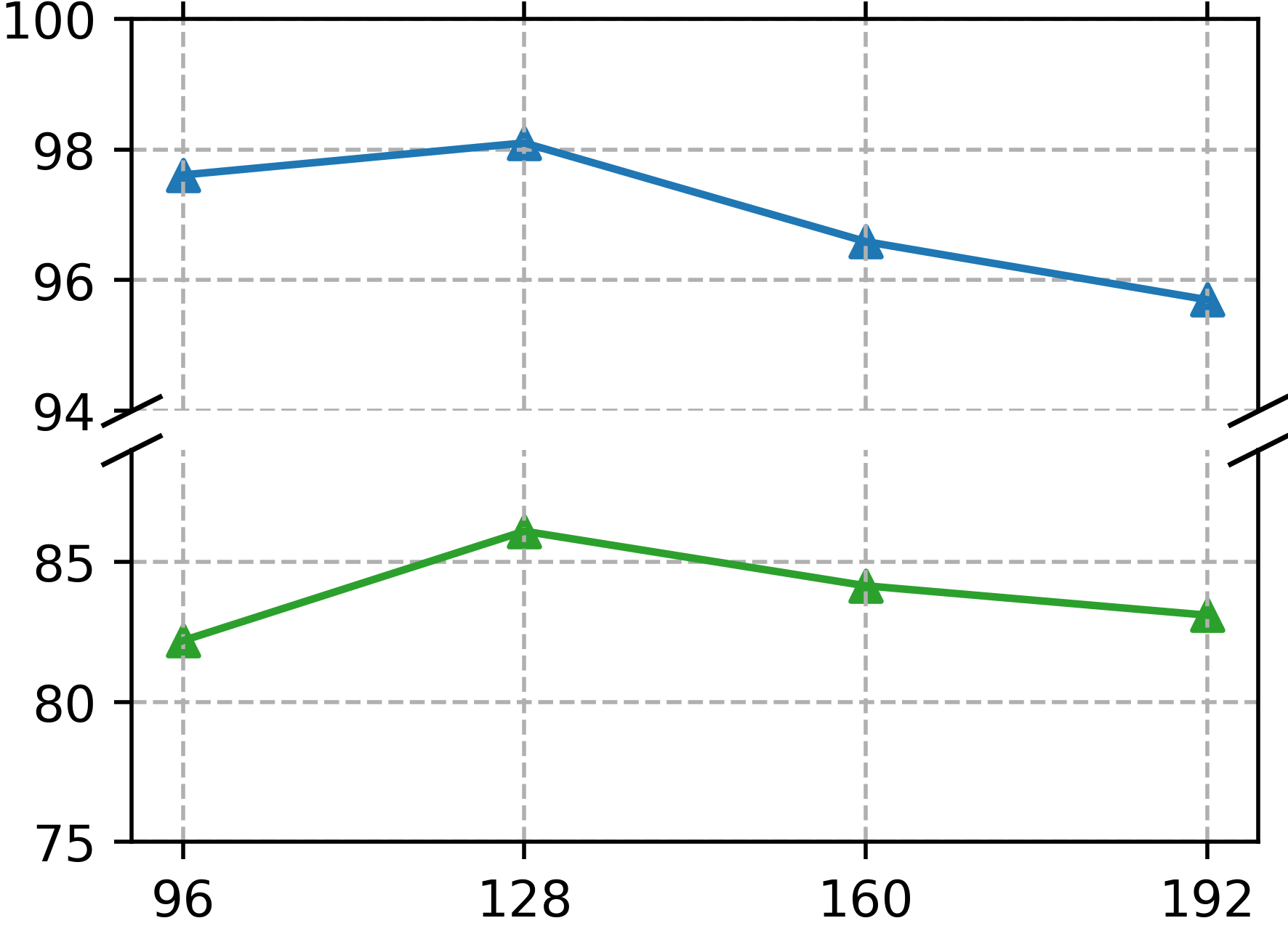}}\hspace{0.15cm}
    
    \subfloat[Number of Self-attention Layers]
    {\label{fig:num_lat_blocks}\includegraphics[width=0.48\linewidth]{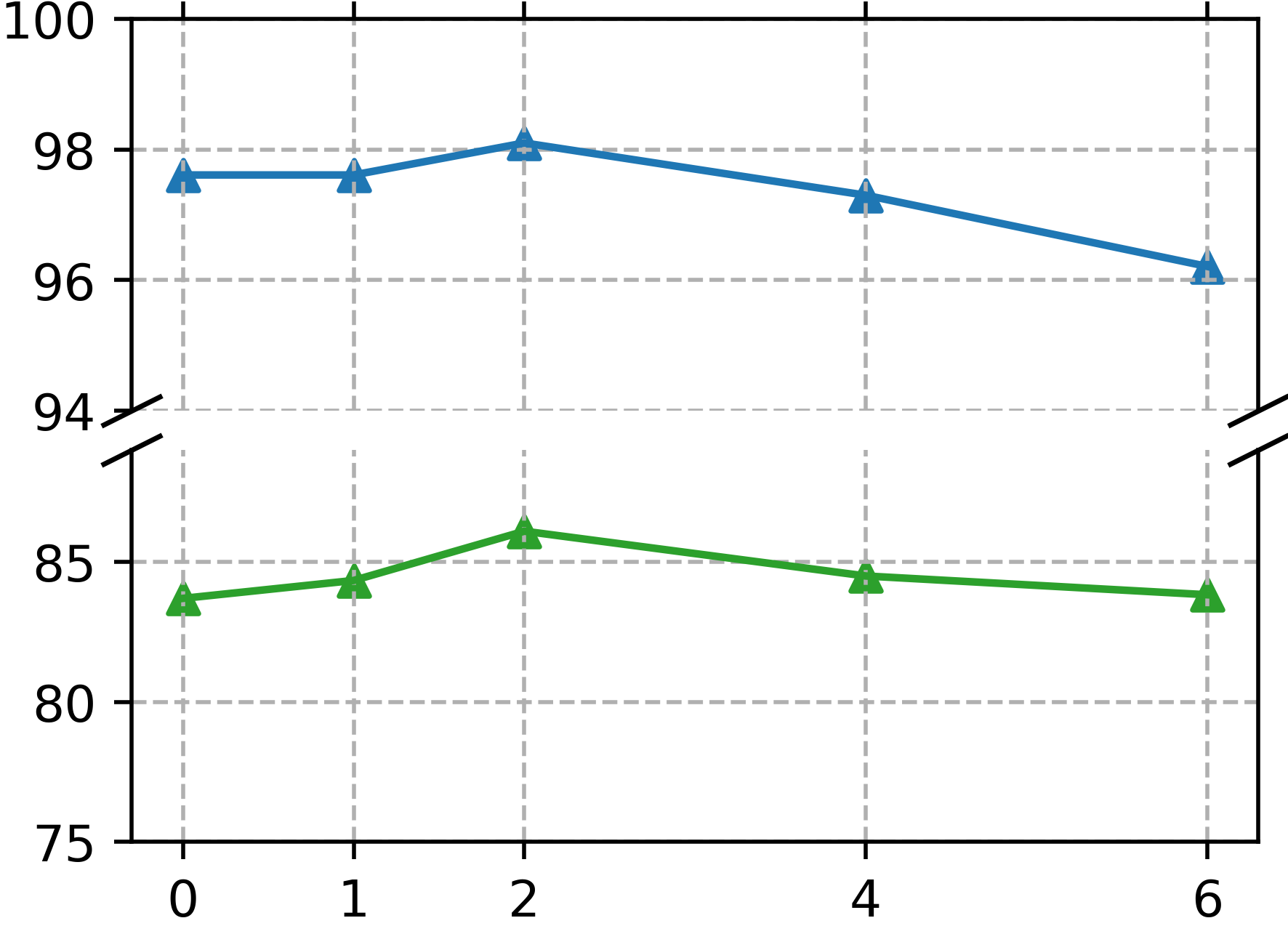}}\hspace{0.15cm}
    \subfloat[Number of Attention Heads]
    {\label{fig:num_heads}\includegraphics[width=0.48\linewidth]{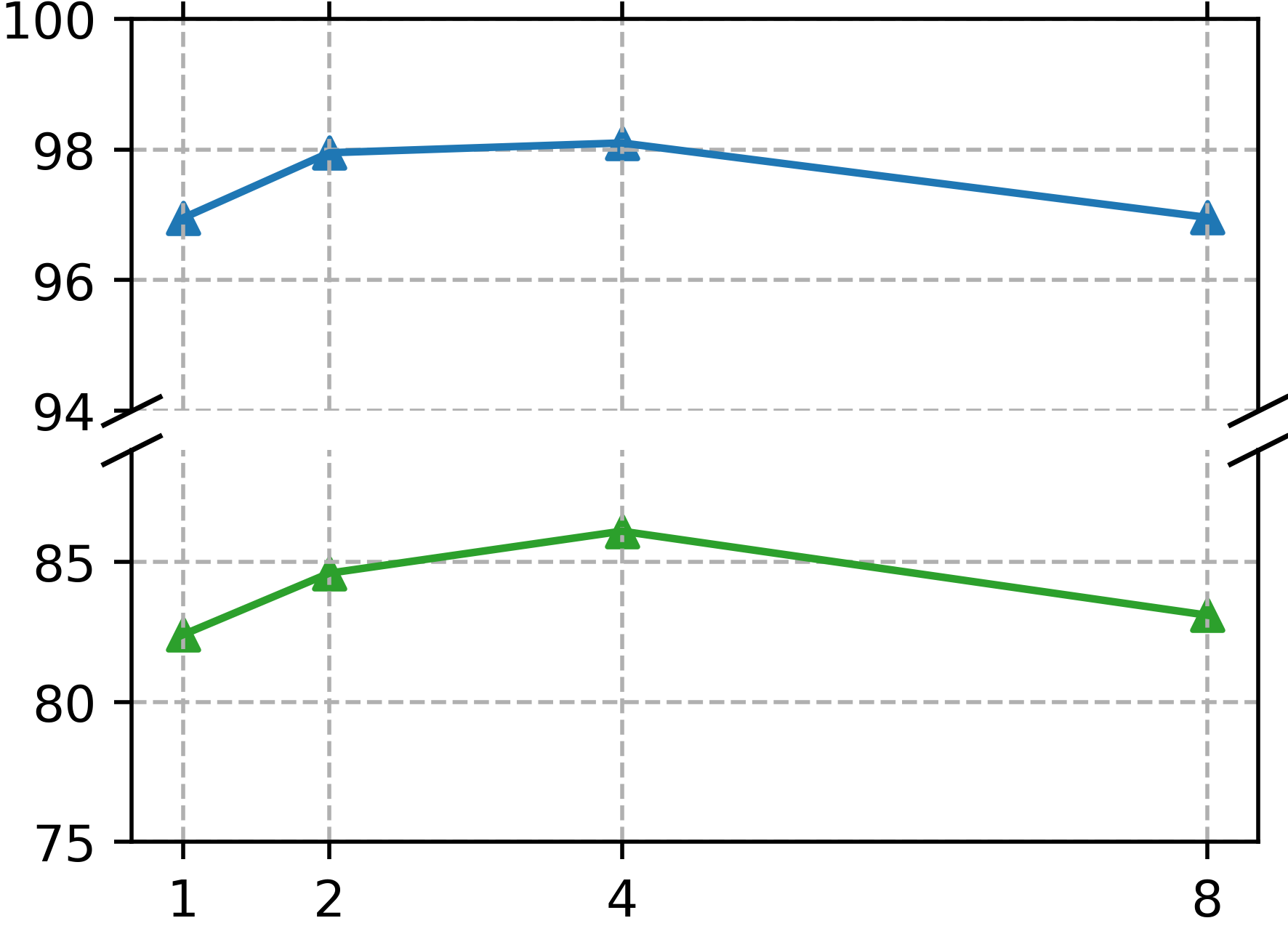}}
    
    \caption{EvT accuracy for different architecture variations using the DVS128 Dataset - 10 classes (top blue line) and the SL-Animals-Dataset - 4 Sets (bottom green line).}
\end{figure}

\paragraph{Transformer hyperparameters.}
The \textit{number of latent vectors} is key for a good generalization.
Figure \ref{fig:num_lat_vec} shows that using too few or too many vectors causes under or over-fitting, leading to suboptimal results.
The \textit{latent vectors dimensionality} (same dimension than the patch tokens) also affects the results. As observed in Fig. \ref{fig:embed_dim}, too short lengths lead to under-fitting results, and too long lengths lead to unstable learning that end with inaccurate results. 

As for the attention hyperparameters, 
Figure \ref{fig:num_lat_blocks} shows how the Self-Attention Fig. allow a better processing of the patch tokens up to a limit where the accuracy is no longer improved.
Similarly, Fig. \ref{fig:num_heads} shows how more attention heads allow to have a finer processing of the input patch tokens, but it can lead to unstable learning and over-fitting.

\paragraph{Attention scores} generated for samples of the used datasets can be found in the supplementary video, and are helpful to understand which data features (patch tokens) lead to classification decisions.
A confusion matrix of the most challenging dataset (SL-Animals) is also provided.

\section{Conclusions}

The present work introduces a novel framework, Event Transformer (EvT), for event data processing.
EvT effectively manages to take advantage of event data properties to minimize its computation and resource requirements, while achieving top-performing results.
EvT introduces a new sparse patch-based event data representation that only accounts for parts of the streams with sufficient logged information, and an efficient and compact transformer-like architecture that naturally processes it.
EvT achieves better or comparable accuracy than the current state-of-the-art on different benchmarks for action and gesture recognition.
Most importantly, our solution requires significantly less computational resources than prior works, being able to perform inferences with minimal latency both in CPU and GPU, facilitating its use on low consumption hardware and real-time applications.

Based on the presented results, we believe that patch-based representations and transformers are a promising line of research for efficient event-data processing. 
This framework also offers promising benefits to other event-based perception tasks, such as 
body tracking or depth estimation,
and to different kinds of sparse data, like LiDAR data.

\section*{Acknowledgments}
This research has been funded by 
FEDER/Ministerio de Ciencia, Innovación y Universidades – Agencia Estatal de Investigación project
PGC2018-098817-A-I00,
DGA T45 17R/FSE and the Office of Naval Research Global project ONRG-NICOP-N62909-19-1-2027.


{\small
\bibliographystyle{ieee_fullname}
\bibliography{egbib}
}

\end{document}